# Lifted Inference for Relational Continuous Models

**Jaesik Choi** and **Eyal Amir**
Computer Science Department
University of Illinois at Urbana-Champaign
Urbana, IL, 61801, USA
{jaesik,eyal}@illinois.edu

**David J. Hill**
Department of Civil and Environmental Engineering
Rutgers, The State University of New Jersey
Piscataway, NJ 08854, USA
ecodavid@rci.rutgers.edu

## Abstract

Relational Continuous Models (RCMs) represent joint probability densities over attributes of objects, when the attributes have continuous domains. With relational representations, they can model joint probability distributions over large numbers of variables compactly in a natural way. This paper presents a new exact lifted inference algorithm for RCMs, thus it scales up to large models of real world applications. The algorithm applies to *Relational Pairwise Models* which are (relational) products of potentials of arity 2. Our algorithm is unique in two ways. First, it substantially improves the efficiency of lifted inference with variables of continuous domains. When a relational model has Gaussian potentials, it takes only linear-time compared to cubic time of previous methods. Second, it is the first exact inference algorithm which handles RCMs in a lifted way. The algorithm is illustrated over an example from econometrics. Experimental results show that our algorithm outperforms both a ground-level inference algorithm and an algorithm built with previously-known lifted methods.

## 1 Introduction

Many real world systems are described by continuous variables and relations among them. Such systems include measurements in environmental-sensors networks (Hill et al., 2009), localizations in robotics (Limketkai et al., 2005), and economic forecastings in finance (Niemira & Saaty, 2004). Once a relational model among variables is given, inference algorithms can solve value prediction problems and classification problems.

At a ground level, inference with a large number of continuous variables is non-trivial. Typically, inference is the task of calculating a marginal over variables of interest. Suppose that a market index has a relationship with $n$ variables, revenues of $n$ banks. When marginalizing out the market index, the marginal is a function of $n$ variables (revenues of banks), thus marginalizing out remaining variables becomes harder. When $n$ grows, the computation becomes expensive. For example, when relations among variables follow Gaussian distributions, the computational complexity of the inference problem is $O(|U|^3)$ ($U$ is a set of ground variables). Thus, the computation with such models is limited to moderate-size models, preventing its use in the many large, real-world applications.

To address these issues, Relational Probabilistic Languages (RPLs) (Ng & Subrahmanian, 1992; Koller & Pfeffer, 1997; Pfeffer et al., 1999; Friedman et al., 1999; Poole, 2003; de Salvo Braz et al., 2005; Richardson & Domingos, 2006; Milch & Russell, 2007; Getoor & Taskar, 2007) describe probability distributions at a relational level with the purpose of capturing larger models. RPLs combine probability theory for handling uncertainty and relational models for representing system structures. Thus, they facilitate construction and learning of probabilistic models for large systems. Recently, (Poole, 2003; de Salvo Braz et al., 2005; Milch et al., 2008; Singla & Domingos, 2008) showed that such models enable more efficient inference than possible with propositional graphical models, when inference occurs directly at the relational level.

Present exact lifted inference algorithms (Poole, 2003; de Salvo Braz et al., 2006; Milch et al., 2008) and those developed in the efforts above are suitable for discrete domains, thus can in theory be applied to continuous domains through discretization. However, the precision of discretizations deteriorates exponentially in the number of dimensions in the model, and the number of dimensions in relational models is the number of ground random variables. Thus, discretization and usage of discrete lifted inference algorithms is highly imprecise.

Here, we propose the first exact lifted inference algorithm for Relational Continuous Models (RCMs), a new relational probabilistic language for continuous domains. Our main insight is that, for some classes of potential functions (or potentials), marginalizing out a ground random variable in a RCM can yield a RCM representation that does not force other random variables to become propositional (Section 4). Further, relational pairwise models, i.e. products of relational potentials of arity 2, remain relational pairwise models after eliminating out ground random variables in those models. Thus, it leads to the compact representations and the efficient computations. We report Gaussian potentials which satisfy the conditions for relational pairwise models (Section 5). However, we are unsure whether the conditions are only satisfied by Gaussian potentials, yet.

We also adapt principles of *Inversion Elimination*, a method devised by (Poole, 2003), to continuous models. *Inversion Elimination*'s step essentially takes advantage of an ability to exchange sums and products. The lifted exchange of sums and products translates directly to continuous domains. This is a unique approach to continuous models, even though the insight is brought from discrete models.

Given a RCM, our algorithm marginalizes continuous variables by analytically integrating out random variables except query variables. It does so by finding a variable, and eliminating it by *Inversion Elimination*. If such elimination is not possible, *Relational Atom Elimination* eliminates each pairwise form in a linear time. If the marginal is not in pairwise form, it converts the marginal into a pairwise form.

This paper is organized as follows. Section 2 provides the formal definition of RCMs. Section 3 overviews our inference algorithms. Section 4 presents main intuitions and results in a Gaussian potential. Section 5 provides the generalized algorithm for relational pairwise models. Section 6 provides experimental results followed by related works in Section 7. We conclude in Section 8.

## 2 Relational Continuous Models

We present a new relational model for continuous variables, Relational Continuous Models (RCMs). Relations among attributes of objects are represented by *Parfactor* models.[1] Each *parfactor* $(L, C, A_R, \phi)$ is composed of a set of logical variables $(L)$[2], constraints on $L$ $(C)$, a list attributes of objects $(A_R)$, and a potential on $A_R$ $(\phi)$. Here, each attribute is a random variable with a continuous domain.

We define a *Relational Atom* to refer the set of ground attributes compactly. For example, *Revenue[B]* is a relational atom which refers to revenues of banks (e.g. $B$ = {'Pacific Bank', 'Central Bank', $\cdots$}). To make the *parfactor* compact, a list of relational atoms is used for

[1] Part of its representation and terms are based on the previous works (Poole, 2003; de Salvo Braz et al., 2005; Milch & Russell, 2007). However, our representaion allows continuous random variables.

[2] Instead of objects, we use the general term, logical variables.

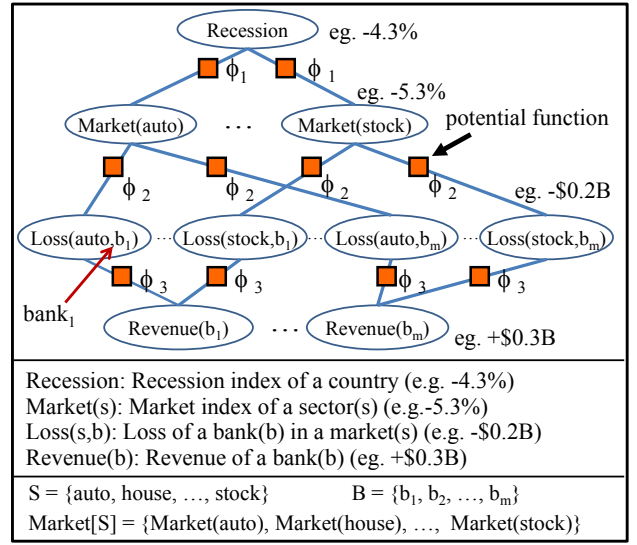

Recession: Recession index of a country (e.g. -4.3%)
Market(s): Market index of a sector(s) (e.g.-5.3%)
Loss(s,b): Loss of a bank(b) in a market(s) (e.g. -$0.2B)
Revenue(b): Revenue of a bank(b) (eg. +$0.3B)

S = {auto, house, …, stock}   B = {$b_1$, $b_2$, …, $b_m$}
Market[S] = {Market(auto), Market(house), …, Market(stock)}

Figure 1: This figure shows a model among banks and market indices. *Recession* is a random variable. *Market[S]*, *Gain[S,B]* and *Revenue[B]* are relational atoms. The variable and atoms have continuous domain $[-\infty, \infty]$. For example, *Market(stock)* is $-5.3\%$, and *Loss(stock,$B_m$)* is $-\$0.2B$.

$A_R$. To refer to an individual random variable, we use a substitution $\theta$. For example, if a substitution ($B$ = 'Pacific Bank') is applied to a relational atom, then the relational atom *Revenue[B]* becomes a ground variable *Revenue*('Pacific Bank').[3] Formally, applying a substitution $\theta$ to a parfactor $g = (L, C, A_R, \phi)$ yields a new parfactor $g\theta = (L', C\theta, A_R\theta, \phi)$, where $L'$ is obtained by renaming the variables in $L$ according to $\theta$. If $\theta$ is a ground substitution, $g\theta$ is a factor. $\Theta_g$ is a set of all substitution for a parfactor $g$. The set of *groundings* of a parfactor $g$ is represented as $gr(g) = \{g\theta : \theta \in \Theta_{gr(L:C)}\}$. We use $RV(X)$ to enumerate the random variables in the relational atom $X$. Formally, $RV(\alpha) = \{\alpha[\theta] : \theta \in gr(L)\}$. $LV(g)$ refers the set of logical variables $(L)$ in $g$.

The joint probability density over random variables is defined by *factors* in a *parfactor*. A *factor* $f$ is composed of $A_g$ and $\phi$. $A_g$ is a list of ground random variables (i.e. $(X_1(\theta), \cdots, X_N(\theta))$). $\phi$ is a *potential* on $A_g$: a function from $range(A_g) = \{range(X_1(\theta)) \times \cdots \times range(X_N(\theta))\}$ to non-negative real numbers. The factor $f$ defines a weighting function on a valuation $(v = (v_1, \cdots, v_m))$: $w_f(v) = \phi(v_1, \cdots, v_m))$. The weighting function for a *parfactor* $F$ is the product of weighting function of all factors, $w_F(v) = \prod_{f \in F} w_f(v)$. When $G$ is a set of *parfactors*, the density is

[3] *Revenue*() refers a random variable. *Revenue*[] refers a relational atom.

the product of all factors in G:

$$w_G(v) = \prod_{g \in G} \prod_{f \in gr(G)} w_f(v). \quad (1)$$

For example, consider the model in Figure 1. $S$ and $B$ in $L$ are two logical variables which represent markets and banks respectively. For example, $S$ can be substituted by a specific market sector (e.g. $S$ = 'stock'). A parfactor $f_1 = (\{Market[S], Gain[S, B]\}, \phi_2)$ is defined over two relational atoms, $Market[S]$ and $Gain[S, B]$. $Market(s)$ (one variable in $Market[S]$) represents the quarterly market change (e.g. $Market(auto)$=−3.1%). $Gain(s, b)$ represents the gain of bank $b$ in the market $s$. Given two values, a potential $\phi_1(Market(s), Gain(s, b))$ provides a numerical value. Given all valuations of random variables, the product of potentials is the probability density.

## 3 Algorithm Overview for RCMs

RCMs model large real-world systems in a compact way. One inference task with such models is to find the conditional density of query variables given observations of some variables.

---

PROCEDURE **FOVE-Continuous**($G$,$Q$)
$G$: parfactors, $Q$: random variables (the query).
  1. If $RV(G) = Q$ return $G$
  2. $G \leftarrow$ **SPLIT**($G$,$Q$)
  3. $E \leftarrow$ **FIND-ELIMINABLE**($G$,$Q$)
  4. $G_E \leftarrow \{g \in G : RV(g) \text{ and } RV(E) \text{ intersect }\}$
  5. $G_{\bar{E}} \leftarrow G \setminus G_E$
  6. $g' \leftarrow$ **ELIMINATE-CONTINUOUS**($G_E$,$E$) (Sections 4 and 5)
  7. $G' \leftarrow \{g'\} \bigcup G_{\bar{E}}$
  8. return **FOVE-Continuous**($G'$,$Q$)

PROCEDURE **ELIMINATE-CONTINUOUS**($G$,$E$)
$G$: parfactors, $E$: a random variable to be eliminated
  1. $g \leftarrow (LV(A_G \setminus E), C_G, A_G \setminus E, \prod_{g \in G} \Phi_g^{\frac{|\Theta_G|}{|\Theta_g|}})$
  2. If ($LV(E)=LV(g)$)
       return **Inversion-Elimination(g,E)**
     Else return **Relational-Atom-Elimination(g,E)**

PROCEDURE **FIND-ELIMINABLE**($G$,$Q$)
$G$: parfactors, $Q$: $\subset RV(G)$ ($G$ is split against $Q$)
  1. For $e$ from $A_G \setminus Q$
       $G_e \leftarrow \{g \in G : RV(g) \text{ and } RV(e) \text{ intersect }\}$
       If $LV(e) = LV(G_e)$ return $e$ (for Inversion-Eliminable)
  2. Choose $e$ from $A_G \setminus Q$
  3. return $e$ (for Relational-Atom-Elimination)

---

Figure 2: FOVE_Continuous (First-Order Variable Elimination with continuous variables) algorithm.

Our inference algorithm, FOVE-Continuous (First-Order Variable Elimination), for RCMs recursively eliminates relational atoms. First, it *splits* (terminology of (Poole, 2003); *shattering* in (de Salvo Braz et al., 2005))[4] relational atoms. The *split* operation makes groundings (e.g. $RV(X)$ $RV(Y)$) of every relational atoms (e.g. $X$ $Y$) disjoint. It introduces observations as observations of groundings of separate relational variables. For example, observing $Market(auto) = 30\%$ creates two separate relational atoms: $Market(auto), Market(M)_{M \neq auto}$. The '$M \neq auto$' then appears in parfactors relating to the latter relational atom. After *split*, FIND-ELIMINABLE finds a relational atom which satisfies conditions for one of the elimination algorithms: *Inversion-Elimination* (Section 5.2) and *Relational-Atom-Elimination* (Section 5.3). The found atom is eliminated by our *ELIMINATE-CONTINUOUS* algorithm explained in Sections 4 and 5. It iterates the elimination until only query variables are remained. The procedure is described in Figure 2.

---

[4] Please refer (Poole, 2003; de Salvo Braz et al., 2005) for further details.

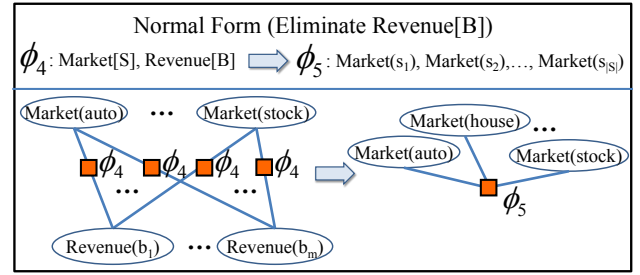

Figure 3: This figure shows a challenging problem in a RCM when eliminating a set of variables (*Revenue[B]*). Eliminating *Revenue[B]* in $\phi_4$ generates an integral $\phi_5$ that makes all variables in *Market[S]* ground. Thus, the elimination makes the RCM into a ground network.

Our main contributions are focused on the algorithm **ELIMINATE-CONTINUOUS**, a lifted variable eliminations for continuous variables. We describe details in Sections 4 and 5.

## 4 Inference with Gaussian Potentials

This section presents our first main technical contribution, efficient variable elimination algorithms for relational Gaussian models. We focus on the inference problem of computing the posterior of query variables given observations. It is important to efficiently integrating out relational atoms (e.g. Revenue[B] = \{$Revenue(b_1)$, $\cdots$, $Revenue(b_m)$\}) for solving this inference problem.

In the following description, we omit the (inequality between logical variables and objects) constraints from parfactors. This allows us to focus on the potential functions inside those parfactors. The treatment below holds with little change for parfactors with such constraints.

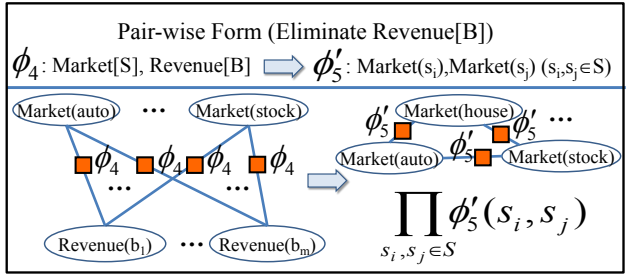

Figure 4: This figure shows our method for the problem shown in Figure 3. When eliminating *Revenue[B]*, we do not generate a ground network. Instead, we directly generate the pairwise form which allows the inference at the lifted level.

## 4.1 Relational Pairwise Potentials

This section focuses on the product of potentials which we call *Relational Normals* (RNs). A RN is the following function with arity 2 (Section 5 provides a generalization for arbitrary potentials).:

$$\phi_{RN}(X,Y) = \prod_{x \in X, y \in Y} \frac{1}{\sigma\sqrt{2\pi}} exp\left(-\frac{(x-y)^2}{2\sigma^2}\right)$$

This potential indicates that the difference between two random variables follows Gaussian distributions.

Consider the models shown in Figure 3 and 4. The models represent the relationships between each market change and revenue of each bank. To simplify notations, we respectively shorten *Market*(s), *Gain*(s, b) and *Revenue*(b) to $M(s)$, $G(s,b)$ and $R(b)$. The potential $\phi_4$ in these figures is $\phi_{RN}(M(s), R(b))$, and the product of potential is $\prod_{s \in S, b \in B} \phi_{RN}(M(s), R(b))$

Figure 4 shows that integrating out a random variable $R(b_i)$ from the joint density results in the product of RNs again (c and c' are constants) as follow.

$$\int_{R(b_i)} \prod_{s \in S} \phi_4(M(s), R(b_i)) = c \cdot exp\left(\frac{(\sum_{s \in S} M(s))^2}{2\sigma^2 \cdot |S|} - \frac{\sum_{s \in S} M(s)^2}{2\sigma^2}\right)$$

$$= c \cdot \prod_{1 \le i < j \le |S|} exp\left(-\frac{(M(s_i) - M(s_j))^2}{2\sigma^2 \cdot |S|}\right) = c' \cdot \prod_{1 \le i < j \le |S|} \phi'_5(M(s_i), M(s_j)) \quad (2)$$

Note that, following equations holds for integration.

$$\int_{R(b_i)} exp\left(-a R(b_i)^2 + 2b R(b_i) + c\right) = \sqrt{\frac{\pi}{a}} exp\left(\frac{b^2}{a} + c\right) \quad (3)$$

Here, the terms *a* and *b* can include random variables except $R(b_i)$.

**Definition 1 (*Connected Relational Normal*)** The product of RNs is connected, when the connectivity graph is a connected component. Each vertex of the connectivity graph is a random variable or a constant in RNs, and each edge is a potential (RN). ∎

**Lemma 1** *The product of RNs is a probability density function when it is connected, and at least a RN includes a constant argument.*

The proof is provided in Section 9.

## 4.2 Constant Time Relational Atom Eliminations

We provide two constant time elimination algorithms for RNs involving a single relational potential $\phi$ (i.e. the product of potentials over different instances of relational atoms). The algorithms eliminate variables, while maintaining the same form, the product of RNs.

### 4.2.1 Elimination of a relational atom $X$ in $\phi_{RN}(X,Y)$

The first problem is to marginalize a relational atom ($X$) in the product of RNs with two relational atoms ($X$, $Y$): $\phi_{RN}(X,Y)$. The potential is the product of $|X| \cdot |Y|$ RNs. Note that each random variable in $X$ has a relation with each variable in $Y$.

**Algorithm 'Pairwise $Constant_1$'**

It marginalizes $x_i$ in $X$, and converts the marginal into a pairwise form.

$$\int_{x_i} \prod_{y \in Y} exp\left(-\frac{(\mathbf{x_i} - \mathbf{y})^2}{2\sigma^2}\right) = \prod_{y_i, y_j \in Y, i < j \le |Y|} exp\left(-\frac{(y_i - y_j)^2}{2\sigma^2 \cdot |Y|}\right) \quad (4)$$

Note that the marginal over $x_i \in X$ and the marginal over $x_j \in X$ ($i \ne j$) are identical. Thus, the following result is derived when it marginalizes all variables in $X$.

$$\int_{x_1} \cdots \int_{x_{|X|}} \prod_{x_i \in X} \prod_{y \in Y} exp\left(-\frac{(\mathbf{x_i} - \mathbf{y})^2}{2\sigma^2}\right)$$

$$= \prod_{x_i \in X} \left(\int_{x_i} \prod_{y \in Y} exp\left(-\frac{(\mathbf{x_i} - \mathbf{y})^2}{2\sigma^2}\right)\right) = \left(\prod_{y_i, y_j \in Y, i < j \le |Y|} exp\left(-\frac{(y_i - y_j)^2}{2\sigma^2 \cdot |Y|}\right)\right)^{|X|}$$

$$= \prod_{y_i, y_j \in Y, i < j \le |Y|} exp\left(-\frac{|X|(y_i - y_j)^2}{2\sigma^2 |Y|}\right) \quad (5)$$

The result of integration is the product of pairwise RNs ($\phi_{RN}(Y,Y)$) with the parameter $\frac{|X|}{2\sigma^2 \cdot |Y|}$.

**Theorem 2** *For the product of RNs between two relational atoms ($\phi_{RN}(X,Y)$), 'Pairwise $Constant_1$' eliminates all ground variables of a relational atom in a constant time.*

**Proof** Eliminating a variables $x_i$ in $X$ takes a constant time shown as Equation 4. Eliminating other variables in $X$ takes a constant time shown as Equation 5. Thus, the computation takes only a constant time without an iteration. ∎

### 4.2.2 Elimination of $n$ random variables in $\phi_{RN}(X,X)$

The second problem is to marginalize some ($n$) variables in a relational atom ($X$) in the product of RNs within the relational atom: $\phi_{RN}(X,X)$. The potential is the product of

$\frac{|X| \cdot (|X|-1)}{2}$ pairwise RNs between two ground random variables in $X$.

**Algorithm 'Pairwise $Constant_2$'**

It updates the marginal after eliminating a random variable without an iteration. When it eliminate $x_m$, it calculates the parameters of $\phi''_{RN}$ given $\phi_{RN}$ as the following equation.

$$\int_{x_m} \prod_{1 \le i < j \le m} \phi_{RN}(x_i, x_j) = \prod_{1 \le i < j \le m-1} \phi_{RN}(x_i, x_j) \cdot \int_{x_m} \prod_{1 \le i \le m-1} \exp\left(-\frac{(x_i - x_m)^2}{2\sigma^2}\right)$$

$$= \prod_{1 \le i < j \le m-1} \phi_{RN}(x_i, x_j) \cdot \prod_{1 \le i \le m-1} \exp\left(-\frac{(x_i - x_j)^2}{2\sigma^2 \cdot (m-1)}\right)$$

$$= \prod_{1 \le i < j \le m-1} \phi_{RN}(x_i, x_j) \cdot \prod_{1 \le i \le m-1} \phi'_{RN}(x_i, x_j) = \prod_{1 \le i < j \le m-1} \phi''_{RN}(x_i, x_j)$$

The coefficient of $\phi''_{RN}$ is the sum of coefficient of $\phi_{RN}$ ($\sigma^2$) and coefficient of $\phi'_{RN}$ ($\sigma^2(m-1)$). The sum of two coefficients results in $\sigma^2 \cdot \frac{m-1}{m}$. Similarly, eliminating the next random variable $\alpha_{m-1}$ results in $\sigma^2 \frac{m-2}{m}$ ($=\sigma^2 \frac{m-1}{m} \frac{m-2}{m-1}$). Thus, eliminating $n$ random variables results in $\sigma^2 \frac{m-n}{m}$ without iterations.

**Theorem 3** *For the product of RNs with a relational atom ($\phi_{RN}(X, X)$), 'Pairwise $Constant_2$' eliminates $n$ ground variables of the relational atom in a constant time.*

**Proof** Updating the parameter of $\phi_{RN}(X, X)$ from $\sigma^2$ to $\sigma^2 \frac{m-n}{m}$ takes only a constant time. ∎

### 4.3 A Linear Time Relational Atom Elimination

This section provides a linear time variable elimination algorithm $O(|U|)$ which can be applied to any product of RNs. This algorithm is used when the constant time algorithms of the previous sections are not applicable.

#### 4.3.1 Elimination of multiple atoms in $\prod \phi_{RN}(X_i, X_j)$

This problem is to marginalize some variables in $U$, ($U = \{X_1, X_2, \cdots, X_{|N|}\}$) in the product of RNs between two relational atoms: $\prod \phi_{RN}(X_i, X_j)$. If all relational atoms have pairwise relationships among each other, there are $\frac{|N| \cdot |N-1|}{2}$ pairwise RNs.

**Lemma 4** *For $|U|$ variables in $|N|$ relational atoms ($U = \{X_1, X_2, \cdots, X_{|N|}\}$) and RN potentials, marginalizing $n$ variables in a ground model takes $O(n \cdot |U|^2)$.*

**Proof** Suppose we eliminate a variable $x \in U$. Eliminating a variable $x$ in RN needs updates coefficients of terms ($x_i x_j$) where $x_i$ and $x_j$ have relations with the variable $x$. When $x$ has relations with all other variables in $U$, the number of terms is bounded by $O(|U|^2)$. Thus, eliminating $n$ variables takes $O(n \cdot |U|^2)$ because it needs $n$ iterations. ∎

Thus, any inference algorithm in a ground model has an order of $O(|U|^3)$ time complexity, when it eliminates all ground variables except a few query variables.

**Algorithm 'Pairwise *Linear*'**

To reduce the time complexity, our lifted algorithm uses following notations which refer ground variables in an atom $X$ compactly: $X_{[m]} = \sum_{1 \le i \le m} x_i$; $X_{[m]^2} = \sum_{1 \le i \le m} x_i^2$; and $X_{[m][m]} = \sum_{1 \le i < j \le m} x_i \cdot x_j$. The notations give the following properties (when $|X| = m$ and $|Y| = n$):

$$(X_{[m]})^2 = X_{[m]^2} + 2X_{[m][m]}$$

$$exp\left(2X_{[m][m]} - (m-1)X_{[m]^2}\right) = \prod_{x_i, x_j \in X} exp\left(-(x_i - x_j)^2\right) = \phi'_{RN}(X, X)$$

$$exp\left(2X_{[m]}Y_{[n]} - nX_{[m]^2} - mY_{[n]^2}\right) = \prod_{x_i \in X, y_k \in Y} exp\left(-(x_i - y_k)^2\right) = \phi''_{RN}(X, Y)$$

For the product of potentials over $X$, $Y$, and $\{x'\}$, our algorithm marginalizes $x'$:

$$\int_{x'} \phi_{RN}(X, x') \cdot \phi_{RN}(Y, x')$$

$$= \int_{x'} exp\left(-(m+n)x'^2 + 2(X_{[m]} + Y_{[n]})x' - (X_{[m]^2} + Y_{[n]^2})\right)$$

$$= \sqrt{\frac{\pi}{m+n}} \cdot exp\left(\frac{(X_{[m]} + Y_{[n]})^2}{m+n} - (X_{[m]^2} + Y_{[n]^2})\right)$$

$$= c \cdot exp\left(\frac{2X_{[m][m]} + 2X_{[m]}B_{[n]} + 2Y_{[n][n]} - (m+n-1)(X_{[m]^2} + Y_{[n]^2})}{m+n}\right)$$

$$= c \cdot \phi'_{RN}(X, X) \cdot \phi''_{RN}(X, Y) \cdot \phi'''_{RN}(Y, Y) \quad (6)$$

It iterates until all $n$ variables are eliminated.

**Theorem 5** *For $|U|$ variables in $|N|$ relational atoms ($U = \{X_1, X_2, \cdots, X_{|N|}\}$) and potentials in RN, 'Pairwise Linear' eliminates $n$ variables in $O(n \cdot |N|^2)$.*

**Proof** WLOG, we marginalize a variable $x' \in X_1$. We make an artificial atom $Y$ which includes all relational atoms, when those atoms have relationships with $X_1$.[5] Then, $\{x'\}$ is split from $X_1$ ($X_1 = X'_1 \cup \{x'\}$ and $X'_1 \cap \{x'\} = \emptyset$). When marginalizing $x'$ out in $\phi_{RN}(X'_1, x') \cdot \phi_{RN}(Y, x')$, the marginal is also the product of RNs shown as Equation 6: $\phi'_{RN}(X'_1, X'_1) \cdot \phi''_{RN}(X'_1, Y) \cdot \phi'''_{RN}(Y, Y)$.

The marginal can be represented without the artificial atom $Y$ in the following procedures. We convert into $\phi''_{RN}(X', Y)$ and $\phi'''_{RN}(Y, Y)$ as follows. First, $\phi''_{RN}(X'_1, Y)$ is represented as the product of RNs between atoms $X_i$ in $Y$ and $X'_1$: $\prod_{X_i \in Y} \phi''_{RN}(X'_1, X_i)$. Second, $\phi'''_{RN}(Y, Y)$ is also represented as the product of RNs between atoms $X_i$ and $X_j$ in $Y$: $\prod_{X_i, X_j \in Y} \phi''_{RN}(X_i, X_j)$.

For each elimination, it updates parameters of all possible pairs $O(|N|^2)$ among $|N|$ atoms. Thus, the computational complexity to eliminate $n$ variables is the order of $O(n \cdot |N|^2)$. ∎

Thus, 'Pairwise *Linear*' has linear time complexity $O(|U|)$ with respect to the number of ground variables.

---
[5]That is, $Y = \bigcup_i X'_i$ and $X'_i = \{\frac{x}{\sigma_i} | x \in X_i\}$, when $\sigma_i$ is the variance used in $\phi_{RN}(X_1, X_i)$.

# 5 Exact Lifted Inference with RCM

This section presents our algorithm, *ELIMINATE-CONTINUOUS*, which generates a new parfactor after eliminating a set of relational atoms given a set of parfactors. A potential of each parfactor is the product of *Relational Pairwise Potentials (RPPs)*:

$$\phi_{RPP}(X,Y) = \prod_{x \in X, y \in Y} \phi_{RPP}(x,y)$$

A *relational pairwise model* is a RCM whose potentials are RPPs. Here, *RPPs* are not limited to the RNs in Section 4.1.

## 5.1 Conditions for Exact Lifted Inference

The lifted *ELIMINATE CONTINUOUS* algorithm provides the exact solution for potentials of parfactors when the potentials satisfy three conditions: *Condition (I)*, analytically integrable; *Condition (II)*, closed under product operations; and *Condition (III)*, closed under marginalizations, thus represented with the product of *relational pairwise potentials* again. The RNs are an example that satisfies the conditions. Here, we introduce another potential, a linear Gaussian, which satisfies the conditions.

**Lemma 6** *The product of RNs with non-zero Means (RNMs) satisfies the three conditions. A RNM has the following form (d is a constant).*

$$\phi_{RN}(X,Y) = \prod_{x \in X, y \in Y} \frac{1}{\sigma\sqrt{2\pi}} exp\left(-\frac{(x-y-d)^2}{2\sigma^2}\right)$$

The proof is provided in Section 9.

## 5.2 Inversion-Elimination

Inversion elimination is applicable when the set of logical variables in $g$ is same with the set of logical variables in $e$, $LV(e) = LV(g)$. Let $\theta_1,...,\theta_n$ be enumeration of $\Theta_g$.

$$\int_{RV(e)} \phi(g) = \int_{RV(e)} \prod_{\theta \in \Theta_g} \phi_g(A_g\theta) = \int_{e[\theta_1]} \cdots \int_{e[\theta_n]} \phi_g(A_g\theta_1) \cdots \phi_g(A_g\theta_n)$$

$$= \prod_{\theta \in \Theta_g} \int_{e[\theta]} \phi_g(A_g\theta)(\because split \text{ (Section 3)}) = \prod_{\theta \in \Theta_g} \int_e \phi_g(A'\theta, e)$$

$$= \prod_{\theta \in \Theta_g} \phi'(A'\theta) = \phi_{g'}$$

Return to the econometric market example, inversion elimination can be applied to $G[S,B]$. Before an elimination, it combines two parfactors which include $\phi_2$ and $\phi_3$ respectively. The combined parfactor is $g = (\{S,B\}, \top, (M[S], G[S,B], R[B]), \phi_2 \cdot \phi_3)$. Then, the elimination procedure is follow.

$$\int_{RV(G)} \phi(g) = \int_{RV(G)} \prod_{s \in S, b \in B} \phi_g(M(s), G(s,b), R(b))$$

$$= \prod_{s \in \{auto,\cdots,stock\}, b \in \{b_1,\cdots,b_m\}} \left(\int_{G(s,b)} \phi_g(M(s), G(s,b), R(b))\right)$$

$$= \prod_{s \in \{auto,\cdots,stock\}, b \in \{b_1,\cdots,b_m\}} \phi_{new}(M(s), R(b)) = \phi_{new}(M[S], R[b]) = \phi_{g'}$$

Note that, the number of substitutions ($|\Theta_g|$) is the number of market sectors ($|S|$) times the number of banks ($|B|$). Regardless the number of substitutions, we can apply the same integration to eliminate $|S| \cdot |B|$ number of random variables (G(s,b)). Thus, it calculates the integral ($= \int_L \phi_g(M(s), G(s,b), R(b))$) one time regardless of specific $s$ and $b$. The marginal ($\phi_{new}(M[S], R[B])$) becomes the potential of the output parfactor ($g'$).

## 5.3 Relational-Atom-Elimination

*Relational-Atom-Elimination* marginalizes atoms when *Inversion-Elimination* is not applicable. It is a generalized algorithm of those for *RN* shown in Section 4. It marginalizes each relational atom of a parfactor $g$ according to three cases: (1) variables in the atom $e$ has relationship with an atom (i.e. '$\phi(X,Y)$'); (2) variables in the atom $e$ has relationships only each other (i.e. '$\phi(X,X)$'); and (3) other general cases (i.e. '$\prod \phi(X_i, X_j)$').

For the case (1), a modified '*Pairwise Constant$_1$*' eliminates an atom $e$. In this case, integrating out a random variable in the atom does not affect integrating another variable in the atom as shown in Section 4.2. That is, $\int_{RV(e)} \prod_{\theta \in \Theta_g} \phi_g(.) = \prod_{\theta_e \in \Theta_e} \int_{e(\theta_e)} \prod_{\theta \in \Theta_{g\setminus\{e\}}} \phi_g(.)$. Here, $E$ is the set of atoms in $g$, and $\bar{E} = E \setminus \{e\}$, and $\Theta_E$ is the set of all substitutions for $E$.

$$\int_{RV(e)} \phi(g) = \int_{RV(e)} \prod_{\theta \in \Theta_E} \phi_g(A_g\theta) = \int_{RV(e)} \prod_{\theta_e \in \Theta_{\{e\}}} \prod_{\theta \in \Theta_{E\setminus\{e\}}} \phi_g(A_g\theta_e, A_g\theta)$$

$$= \prod_{\theta_e \in \Theta_{\{e\}}} \int_{e[\theta_e]} \prod_{\theta \in \Theta_{E\setminus\{e\}}} \phi_g(A_g\theta_e, A_g\theta) = \prod_{\theta_e \in \Theta_{\{e\}}} \phi'(RV(\bar{E}))(\because Condition(I))$$

$$= \phi'(RV(\bar{E}))^{|RV(e)|} = \phi''(RV(\bar{E}))(\because Condition(II))$$

Normally, the marginal $\phi''(RV(\bar{E}))$ is not a *relational pairwise potential* because all random variables in $\bar{E}$ are arguments of the potential. However, when *Condition (III)* is satisfied, the marginal can be converted into the product of *relational pairwise potentials*: $\phi''(RV(\bar{E})) = \prod_{X_i, X_j \in RV(\bar{E})} \phi_{RPP}(X_i, X_j)$.

In the financial example, it eliminates $R[B]$ as follow.

$$\int_{RV(R)} \phi(g') = \int_{RV(R)} \prod_{s \in S, b \in B} \phi_{new}(M(s), R(b))$$

$$= \prod_{b \in B} \int_{R(b)} \prod_{s \in S} \phi_{new}(M(s), R(b)) = \prod_{b \in B} \phi'_{new}(M(auto), \cdots, M(stock))$$

$$= \phi'_{new}(M(auto), \cdots, M(stock))^{|RV(R)|} = \phi''_{new}(M(auto), \cdots, M(stock))$$

Beyond Relational Gaussian defined in Section 4.1, any potential function satisfying the Condition III) can convert the potential $\phi''_{new}$ into the pairwise form $\prod \phi'''_{new}$.

$$\phi''_{new}(M(auto), \cdots, M(stock)) = \prod_{s_1, s_2 \in S} \phi'''_{new}(M(s_1), M(s_2))$$

Likewise, for the cases (2) and (3), generalized algorithms of '*Pairwise Constant$_2$*' and '*Pairwise Linear*' are also applied respectively.

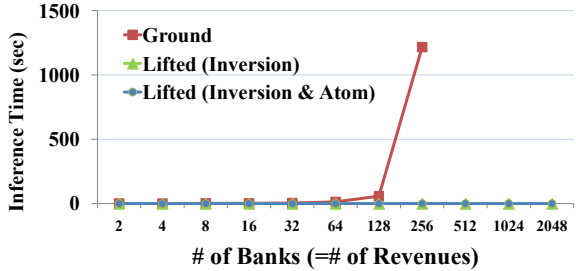

Figure 5: Inference time with different number of banks

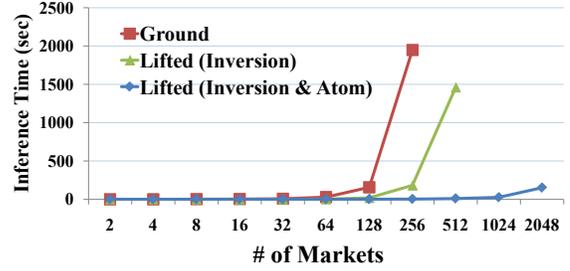

Figure 6: Inference time with different number of markets

## 6 Experiments

We report experiments for the recession model provided in the paper. For experiments, we implemented three algorithms: (A) inference with a grounded model; (B) inference with only Inversion-Elimination; and (C) inference with both Inversion-Elimination and Relational-Atom-Elimination. Our new algorithm (C) is significantly faster than the grounded model (A) and Inversion-Elimination (B). Note that Inversion-Elimination (B) is also our new algorithm for continuous variables, even though comparable elimination methods for discrete variables (de Salvo Braz et al., 2005; Milch et al., 2008; Pfeffer et al., 1999) existed prior to ours. Our experimental results are shown in Figure 5 and 6

In the recession model, we provided observations for one market variable (M) and one revenue variable (R).[6] Those variables were split from relational atoms. Then, we calculated the marginal density of the Recession variable. We increased the number of markets and the number of banks from 2 to 2048 exponentially. We set an hour of cut-off time. With 512 banks, the grounded inference (A) did not complete within an hour. Meanwhile, the Inversion Elimination (B) and our new algorithm (C) finished computations in almost a constant time even for 2048 banks. With 512 markets, (A) could not finish within an hour, again. With 1024 markets, (B) did not finish in an hour. Meanwhile, our new algorithm (C) finished in a reasonable time (about 151 secs) even with 2048 markets.

## 7 Related Work

(Poole, 2003), solves inference problems with the unification which dynamically splits a set of ground nodes and unifies them. With a counting formula, (de Salvo Braz et al., 2005; de Salvo Braz et al., 2006) provides a tractable algorithm. (Milch et al., 2008) applies the counting formula to reduce the size of probability density tables. However, these lifted inference algorithms are hard to apply to continuous domains.

---

[6] Observations are required to make the product of RNs a probability density function. Please refer Lemma 1 for details.

MLNs (Markov Logic Network) (Richardson & Domingos, 2006) use First-order logic sentences to represent relationships over nodes in a graphical model. In this regard, MLNs also represent graphical models at the relational level. (Singla & Domingos, 2008) provides an approximated lifted inference algorithm over discrete domain. (Singla & Domingos, 2007) makes an analysis for infinitely many discrete variables. However, these achievements are not for continuous domains, too. Although there is an inference algorithm for Hybrid MLNs (Wang & Domingos, 2008), it is an approximated algorithm. Thus, most of achievements are comparable to lifted inferences (de Salvo Braz et al., 2005; Milch et al., 2008; Pfeffer et al., 1999) over discrete domain.

Inference with Gaussian distributions is a traditional problem (Roweis & Ghahramani, 1999). In detail, calculating conditional densities of multivariate Gaussians requires matrix inversions (Kotz et al., 2000) which are intractable for high dimensions. (Lerner & Parr, 2001; Shenoy, 2006) builds inference algorithms for hybrid models with Gaussians. (Paskin, 2003) shows that efficient inference is possible for a linear Gaussian when the treewidth of the model is small. For models with large treewidth, however, those inference algorithms over ground models which would be inefficient.

Recent advances in inference with relational models (Kisynski & Poole, 2009; Mihalkova & Mooney, 2009) show the promise of the approach in discrete models, and underline the promise of our algorithm in continuous models.

## 8 Conclusion and Future work

In this paper, we propose a new exact lifted inference algorithm for Relational Continuous Models (RCMs). This algorithm is an advancement of exact inference in RCMs, since all previous works are restricted to discrete domain. Given a query and observations, our algorithm exactly computes the conditional density of the query, when potentials satisfy specified conditions.

There are two limitations in our current algorithm. First,

found potentials which satisfy the conditions in Section 5 are variants of Gaussian potentials. Thus, finding potentials beyond Gaussian is a goal of our future works. Second, the current algorithm is designed only for continuous variables. Many real-world models require not only continuous variables but also discrete variables. Thus, making an efficient inference algorithm for hybrid relational models would be a promising direction.

### Acknowledgements

We wish to thank Abner Guzman Rivera and the anonymous reviewers for their valuable comments. This material is supported by NSF IIS 05-46663 and UIUC/NCSA AESIS 251024 grants.

## 9 Appendix

**Proof of Lemma 1** Here, we prove that the product of RNs integrates to a constant given the conditions. The constant becomes the normalizing factor of the probability density function.

We prove this by contradiction. Suppose that the product of RNs does not integrate to a constant. That is, it integrates to infinity.

According to Equation 2, the product of RNs maintains the same form after integrating out a random variable $x$. Thus, only possible case to be infinity is when the marginal (after an integration over $x$) is a constant function of another random variable $y$ which is not yet integrated.

When $x$ has relations with more than one variable (e.g. $y$ and $z$), the condition for infinity is not satisfied. The marginal includes a potential $\phi(y, z)$. When $x$ has a relation with only $y$ which has relations with other variables beyond $x$, the condition for infinity is not satisfied. The marginal is not a constant function of $y$.

Thus, only $\phi(x, y)$ satisfies the condtion for infinity. Given the assumption that at least a RN includes a constant, $y$ can not be a variable. Thus, it contradicts the assumption. ∎

**Proof of Lemma 6** First, the product of RNMs is analytically integrated by the rule in Equation 3. Thus, the product of RNMs satisfy the *Condition (I)*.

Second, the product of RNMs is closed under product operations and marginalizations. It satisfies the *Condition (I)* because it is an exponential family. That is, the product of two RNMs ($\phi'_{RNM}(x, y)$ and $\phi''_{RNM}(x, y)$) is another RNMs ($\phi'''_{RNM}(x, y)$). Thus, the product of RNMs satisfies the *Condition (II)*.

Third, it is also closed under marginalizations. When $y_j$ in Equation 4 is substituted with $y_j - d$, the following equation is derived.

$$\int_{x_i} \prod_{y \in Y} \exp\left(-\frac{(\mathbf{x_i} - \mathbf{y} - \mathbf{d})^2}{2\sigma^2}\right) = \prod_{y_i, y_j \in Y} \exp\left(-\frac{(y_i - y_j - 0)^2}{2\sigma^2 \cdot |Y|}\right)$$

Thus, the result is the product of RNMs.

As explained in the proof of Theorem 5, the the product of RNMs can be represented as the following form $\phi_{RNM}(X, x') \cdot \phi_{RNM}(Y, x')$ when $x'$ is the variable of integration.

When $y_j \in Y$ in Equation 6 is substituted with $y_j - d \in Y'$, the following equation is derived.

$$\begin{aligned}
\int_{x'} \phi_{RNM}(X, x') \cdot \phi_{RNM}(Y, x') &= c \cdot \int_{x'} \phi_{RN}(X, x') \cdot \phi_{RN}(Y', x') \\
&= c' \cdot \phi'_{RN}(X, X) \cdot \phi''_{RN}(X, Y') \cdot \phi'''_{RN}(Y', Y') \\
&= c'' \cdot \phi'_{RNM}(X, X) \cdot \phi''_{RNM}(X, Y) \cdot \phi'''_{RNM}(Y, Y)
\end{aligned}$$

The result is also the product of RNMs. Thus, it is closed under marginalizations. ∎